\pdfoutput=1

\documentclass[11pt]{article}

\usepackage[final]{acl}

\usepackage{times}
\usepackage{latexsym}

\usepackage[T1]{fontenc}

\usepackage[utf8]{inputenc}

\usepackage{microtype}

\usepackage{inconsolata}

\usepackage{graphicx}
\usepackage{booktabs}
\usepackage{enumitem} 
\usepackage{amsmath} 
\usepackage{url}
\usepackage{hyperref}
\usepackage{tcolorbox}

\title{Thai Winograd Schemas: A Benchmark for Thai Commonsense Reasoning}

\author{Phakphum Artkaew \\
  Department of Electrical and Computer Engineering \\
Tandon School of Engineering \\
 New York University \\
  \texttt{pa2497@nyu.edu}}

\begin{document}
\maketitle
\begin{abstract}
Commonsense reasoning is one of the important aspects of natural language understanding, with several benchmarks developed to evaluate it. However, only a few of these benchmarks are available in languages other than English. Developing parallel benchmarks facilitates cross-lingual evaluation, enabling a better understanding of different languages. This research introduces a collection of Winograd Schemas in Thai, a novel dataset designed to evaluate commonsense reasoning capabilities in the context of the Thai language. Through a methodology involving native speakers, professional translators, and thorough validation, the schemas aim to closely reflect Thai language nuances, idioms, and cultural references while maintaining ambiguity and commonsense challenges. We evaluate the performance of popular large language models on this benchmark, revealing their strengths, limitations, and providing insights into the current state-of-the-art. Results indicate that while models like GPT-4 and Claude-3-Opus achieve high accuracy in English, their performance significantly drops in Thai, highlighting the need for further advancements in multilingual commonsense reasoning.
\end{abstract}

\section{Introduction}
\begin{figure*}[htbp]
    \centering
    \includegraphics[width=\textwidth]{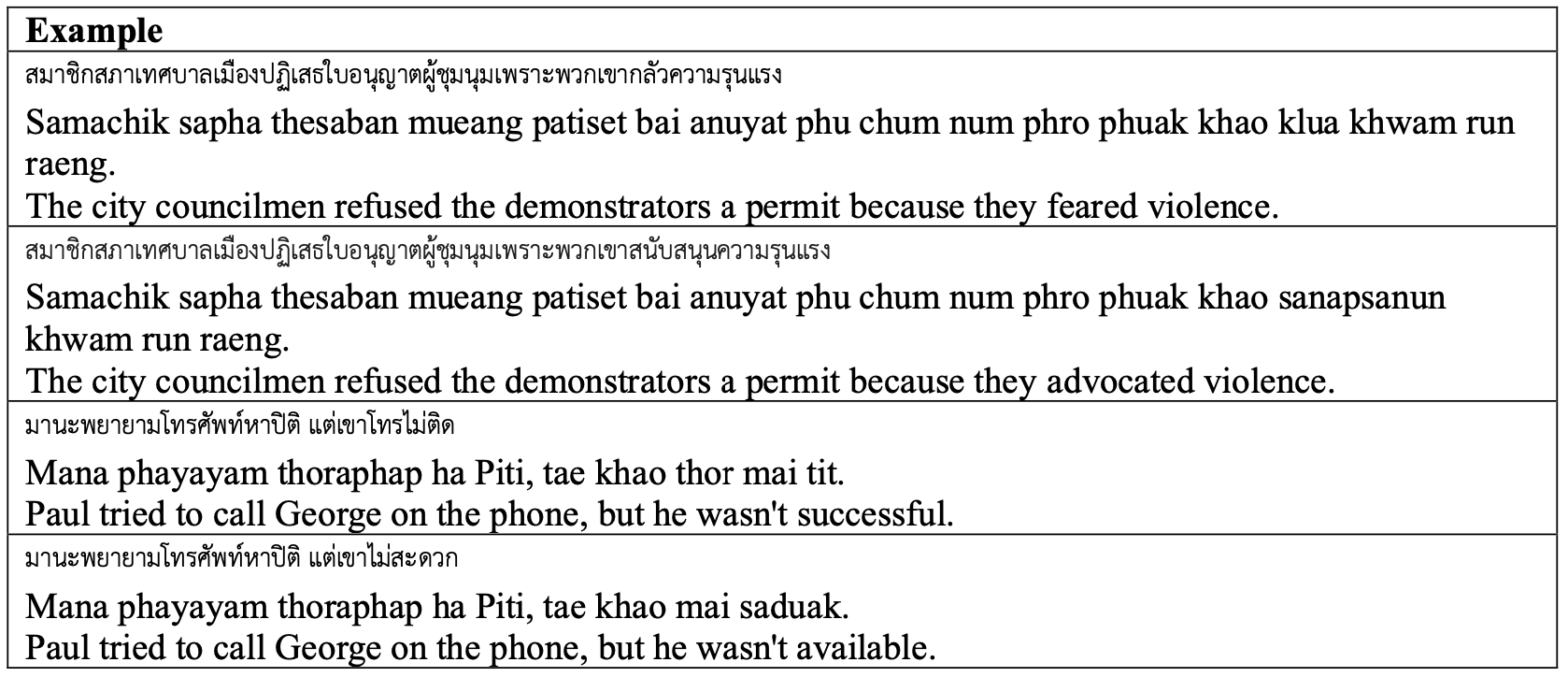}
    \caption{Winograd Schema examples in Thai, with transliteration and corresponding English version.}
    \label{fig:example}
\end{figure*}
Commonsense reasoning is an important challenge in artificial intelligence; however, most resources and benchmarks are in English, with only a few translations \cite{davis2023benchmarks}. The Winograd Schemas Challenge (WSC) has emerged as a widely adopted benchmark for evaluating the commonsense reasoning capabilities of language models \cite{levesque2012winograd}. WSC is featured in both GLUE and SuperGLUE \cite{wang-etal-2018-glue, wang2019superglue}. These benchmarks are widely used to evaluate a model’s understanding of general language. Significant progress has been made in developing commonsense reasoning models for high-resource languages like English. However, their performance on low-resource languages remains largely unexplored. This research gap hinders the development and fair evaluation of multilingual and cross-lingual NLU systems. \citet{linzen-2020-accelerate} argues for developing multilingual training data and benchmarks to avoid English-centric model development, suggesting composite scores that average performance across languages to better evaluate systems.

There have been several one-to-one translations of the Winograd Schemas into various languages, including French \cite{amsili2017google}, Portuguese \cite{de2019winograd}, Mandarin \cite{bernard-han-2020-mandarinograd}, Hebrew \cite{shwartzWinograd}, Hungarian \cite{vadasz2022winograd}, and Russian \cite{shavrina2020russiansuperglue}. There are also various Japanese translations and adaptations \cite{shibata2015nihongo, tanaka2024translations, tanaka2024translationpreserving}. Before our translation, no Thai version of the Winograd Schemas existed. This lack prompted us to take the initiative.

The key contributions of our work are:

\begin{enumerate}
    \item We introduce Thai-WS, the first benchmark dataset for evaluating the Winograd task and commonsense reasoning capabilities of language models and NLP systems in the Thai language.
    \item We evaluate the performance of state-of-the-art models, including GPT-4, GPT-3.5, Claude-3-Haiku, Claude-3-Sonnet, Claude-3-Opus, Typhoon, and Command R+ on the Thai-WS dataset to assess their cross-lingual reasoning abilities. 
\end{enumerate}

\section{Related Work}
\subsection*{Background}
The Winograd Schema Challenge, introduced by \citet{levesque2012winograd} in 2012 as an alternative Turing Test, consists of pairs of similar sentences with slight differences that introduce ambiguities resolved through general knowledge and logical reasoning. A classic example by \citet{winograd1972understanding} illustrates the importance of context: "The city councilmen refused the demonstrators a permit because they feared/advocated violence." The choice of "feared" or "advocated" shifts the referent of "they" from the city councilmen to the demonstrators. This is because "feared" aligns with concerns typical of councilmen's reasons for refusal, whereas "advocated" suggests demonstrators' motivations, reflecting how context dictates meaning. 
\subsection*{Other Thai Reasoning Benchmarks}
To the best of our knowledge, three public benchmarks are available for Thai reasoning. First is XNLI, which is a translation of the MNLI benchmark into Thai, among 14 other languages \cite{conneau2018xnli}. Second is XCOPA, translated from the COPA benchmark to assess causal commonsense reasoning \cite{ponti-etal-2020-xcopa}. Third, the recently released M3Exam, is based on examination questions \cite{zhang2024m3exam}. Unlike these benchmarks, the Winograd Schema Challenge is specifically designed to test commonsense reasoning through pronoun reference disambiguation, a task that requires resolving ambiguity using implicit world knowledge. This makes it fundamentally different and more challenging than tasks like XCOPA, which focus on causal reasoning but do not require the same level of subtle language understanding or contextual interpretation.
\subsection*{A Simple Method for Commonsense Reasoning}
\citet{trinh2018simple} introduced a simple approach to solve the Winograd Schema challenge by substituting one of the pronouns in the sentence and allowing the language model to determine which substitution has a higher probability. This method effectively frames Winograd Schemas as a binary classification task. Since then, many language models, such as GPT-2 \cite{radford2019language} and GPT-3 \cite{brown2020language}, have used this method to evaluate their performance on Winograd Schemas. However, due to limited access to models and computational power, the evaluation of this version of the Winograd Schema Challenge was conducted differently, using a prompt-based method. Details of this method are provided in the experimental setup section.
\section{Dataset Construction}
\subsection{Translation}
Two professional translators, who were native Thai speakers fluent in English and had experience translating from English to Thai, were hired. A one-to-one translation approach was followed to ensure that each schema was translated directly and accurately while preserving its original meaning. In a pilot translation phase, one native speaker translated the first 85 Winograd Schemas. Based on a qualitative analysis of these initial translations, guidelines were provided for a second native speaker to translate the remaining 200 schemas. In total, 285 Winograd Schemas were translated from English to Thai.

Translation guidelines were provided, instructing them to adapt names and contexts to suit the Thai language while preserving the ambiguity and nuances of the original schema. We also note that Thai pronouns are mostly similar to English but with a wider variety of formality levels. For example, in English, addressing an older individual with the pronoun "you" is generally acceptable. However, in Thai, maintaining politeness requires using a specific term of respect when referring to an older person. Therefore, we chose the translation based on the context of the sentence. The translators were also asked to mark any translated names and translations they were unsure about in red, so that the validator in the next step could pay extra attention to those instances. For example, in \autoref{fig:example}, the names Paul and George were changed to Mana and Piti, respectively, adapting the names to better suit the Thai context while preserving the essence of the original content.

Some phrases could be directly translated, while others required adjustments to names and contextual adaptations to better suit the Thai language and culture. However, there were a few instances where the translators highlighted certain phrases in red, indicating that they are worth mentioning. The following includes two examples of schemas highlighted in red and one instance of a word adaptation made to better fit the nuances of the Thai language.

\begin{enumerate}[label=(\arabic*), ref=\arabic*]
    \item \label{1a}
        \begin{enumerate}[label=(\roman*)]
            \item Many people start to read Paul's books and can't put them down. They are gripped because Paul writes so well.
            \item Many people start to read Paul's books and can't put them down. They are popular because Paul writes so well.
        \end{enumerate}
    \item \label{2a}
        \begin{enumerate}[label=(\roman*)]
            \item During a game of tag, Ethan chased Luke because he was "it".
            \item During a game of tag, Ethan ran from Luke because he was "it".
        \end{enumerate}
    \item \label{3a}
        \begin{enumerate}[label=(\roman*)]
            \item Bob was playing cards with Adam and was way ahead. If Adam hadn't had a sudden run of good luck, he would have won.
            \item Bob was playing cards with Adam and was way ahead. If Adam hadn't had a sudden run of good luck, he would have lost.
        \end{enumerate}
\end{enumerate}
Most of the translation problems encountered involve examples like (\ref{1a}), where in English the word "they" can refer to both people and objects. In this case, "they" can mean either "people" or "Paul's books." However, in Thai, it is uncommon to use the same pronoun for both objects and people. Therefore, an alternative pronoun that is acceptable in Thai was chosen. In this instance, the English equivalent word to "those" was used instead.

In some cases, the meaning had to be adjusted to sound more natural in Thai by replacing idioms or figurative language. For example, in Thai, the phrase "he was" in (\ref{2a}) is sufficient to imply that he was the chaser in a game of tag, so the subject "it" is omitted from the sentence.

Throughout the translation process, the translators and validators prioritized making the text sound natural in Thai, which required word adaptations. Instead of strictly adhering to literal translations, wording was selected to better fit Thai language norms and context. For instance, in (\ref{3a}), while a direct translation of “good luck” is understandable to Thai speakers, the context of Bob and Adam playing cards made the Thai equivalent of “hand up” more suitable. Although “good luck” would be acceptable, “hand up” aligns more closely with the situation and feels more natural in Thai. Thus, “hand up” was chosen to maintain contextual and linguistic appropriateness.
\subsection{Validation}
\begin{table*}[htbp]
\centering
\begin{tabular}{lccc}
\hline
Model         & Accuracy (English) & Accuracy (Thai) & Accuracy (Thai Exact) \\ \hline
Typhoon       & 58.60\%               & 56.14\%             & 53.33\% \\
Claude-3-Haiku        & 64.21\%               & 53.33\%             & 52.28\% \\
Claude-3-Sonnet       & 80.70\%            & 66.67\%           & 65.96\% \\
Claude-3-Opus & 92.63\%                & \textbf{79.65\%}             & \textbf{84.21\%} \\
GPT-3.5       & 70.88\%                & 53.33\%             & 52.28\% \\
GPT-4         & \textbf{94.04\%}             & 76.49\%             & 79.65\% \\ 
Command-r-plus& 87.02\%           & 61.75\%             & 64.91\% \\
\hline
Human         & 92\%               & 88\%             & - \\ \hline
\end{tabular}
\caption{Accuracy vs. Model in English, Thai, and Thai Exact}
\label{tab:modelandaccuracy}
\end{table*}

The translated Winograd Schemas were reviewed by three native Thai speakers, and a validator was tasked with identifying potential issues, focusing on text flagged by the translators. Based on their feedback, final adjustments and typographical corrections were made. The dataset is publicly available\footnote{\url{https://huggingface.co/datasets/pakphum/winograd_th}} and consists of a test set containing 285 schemas, each with corresponding choices and answers.

\subsection{Human baseline}
The study for the Thai Winograd Schemas human baseline was conducted in a manner similar to that of \citet{davis2016human}. A total of 30 native Thai speakers participated as volunteers and were divided into two groups of 15. The Winograd Schemas were split into two parts: one group completed part A, while the other group completed part B. Unlike \citet{davis2016human}, who conducted their study in person, this study was conducted virtually using Google Forms. Participants were provided with a link to complete their assigned part of the Winograd Schemas and were not restricted by time limits, allowing them to pause and resume the task as needed. This approach was designed to ensure participants could take their time and complete the study effectively. The observed score was approximately 88\%, representing the average accuracy score of all participants.

\section{Experimental Setup}
Large language models were evaluated on Thai and English Winograd Schemas to assess their Thai language understanding and enable cross-linguistic comparisons of commonsense reasoning. The Winograd Schema Challenge comes in two versions: WSC-273, which contains 273 questions, and WSC-285, an extended version with 12 additional questions, totaling 285 \cite{kocijan2023defeat}. In this study, the WSC-285 dataset was used.

In addition to the main Thai dataset, the models were evaluated on the Thai-exact dataset, which consists of Winograd Schemas translated into Thai using Google Translate, with hand corrections by the authors to address translation mistakes. This evaluation aimed to compare the effect of adapting Thai names and contexts, as done in the main Thai dataset, against the more literal translations in Thai-exact.

The models chosen for evaluation were Typhoon \cite{pipatanakul2023typhoon}, Claude-3 \cite{claude3}, GPT-3.5 \cite{brown2020language} and GPT-4 \cite{achiam2023gpt}, and C4AI Command R+ \cite{cohere2024commandrplus}. Typhoon is a Thai-specific model designed to handle Thai language tasks. Both GPT-3.5 and GPT-4 represent state-of-the-art general-purpose language models, with GPT-4 offering improved reasoning and comprehension capabilities over its predecessor. Claude-3 is another cutting-edge general-purpose language model excelling in similar tasks. C4AI Command R+, on the other hand, is a multilingual model designed to support many languages. This selection was made to evaluate performance across Thai-specific, multilingual, and general-purpose capabilities. Further details about the models can be found in Appendix \ref{sec:testedmodels}.
\subsection*{Implementation and evaluation}
Language models were evaluated using a prompt structure approach. A system prompt and user prompt structure were utilized to assess the models. A system prompt provides context and instructions to language models before a task, specifying the model’s role, personality, tone, or other relevant information to enhance its responses \cite{anthropicSystemPrompts}. The system prompt was designed to prepare the model for the Winograd Schemas task, instructing it to respond with the correct answer. 

Evaluation was conducted by calculating accuracy, with only responses that exactly matched the correct answers in the schemas considered correct. A manual review was performed to address cases where correct answers were presented with additional text, such as “the answer is the city councilmen” instead of “the city councilmen.” This correction process was only required for Typhoon, as it occasionally included additional information in its responses. The models were evaluated via their respective APIs. Further details on the evaluation process and prompt setup are provided in Appendix \ref{sec:prompteval}.
\section{Result and Discussion}
\autoref{tab:modelandaccuracy} presents the accuracy results for each evaluated model. Several human baselines exist for the English Winograd Schema Challenge \cite{kocijan2023defeat}. Our English human baseline is derived from the study conducted by \citet{davis2016human}, which involved human participants evaluating Winograd Schemas in English. In \citet{davis2016human} research, human accuracy was observed to be approximately 92\%.
\subsection*{Comparison of models}
GPT-4 leads with the highest accuracy in English at 94.04\%, while Claude-3-Opus has the best performance in Thai Exact at 84.21\%. In the Thai language context, Claude-3-Opus again performs the best with 79.65\%, followed closely by GPT-4 at 76.49\%. Other models, such as GPT-3.5 and Claude-3-Haiku, show significantly lower accuracy, particularly in the Thai and Thai Exact columns, indicating a drop in performance when these models are applied to languages other than English. Human performance is slightly lower in Thai, but the gap is larger for language models, highlighting challenges in language adaptation.

The performance of the Thai dataset versus the Thai-Exact dataset is relatively similar, with less than a five percent difference across all models. This may suggest that adapting Thai names and nuances does not have a significant impact on model performance.
Overall, the results indicate that all models perform better on the English version of the dataset compared to the Thai versions, whether adapted by humans or translated by machines. The performance drop in Thai suggests that these models struggle more with the linguistic nuances and structures of the Thai language. The performance drop may also be due to data leakage \cite{elazar-etal-2021-back}, which could affect the integrity of the results.
\subsection*{Do larger models consistently perform better}
The analysis of model parameter sizes for GPT and Claude is derived from the assumptions provided by \citet{pmlr-v235-coda-forno24a}. According to their study, GPT-4 has approximately 1760 billion parameters, while Claude-3-opus has around 300 billion parameters. The results indicate that larger models tend to perform better on this task, with performance consistently improving as model size increases. Specifically, the smallest model in the analysis, Typhoon (based on LLaMA 8B), demonstrated the lowest performance. Performance improved with larger models, starting from Command-r-plus (104B) and continuing to larger models like Claude and GPT. However, in the context of Thai, larger models do not always perform better, as Claude-3-Opus outperformed the larger GPT-4. This finding suggests that while model size is a contributing factor, other elements, such as training data quality, multilingual capabilities, or architectural design, may also significantly impact performance.
\subsection*{Are the mistakes similar to those made by English LLMs?}
To understand whether the errors stem from language understanding or commonsense reasoning, we further analyze the models’ output by examining the overlapping incorrect questions in both the English and Thai sets. By observing the percentage of consistently incorrect overlapping questions from the English set to the Thai set (i.e., the number of overlapping incorrect answer in both sets divided by the number of incorrect answer in the English set), we find that the percentage usually exceeds half of the incorrect questions in the Thai set for models such as Typhoon, GPT-3.5, Claude-3-haiku, Claude-3-sonnet, and command-r-plus. This suggests that most models struggle with commonsense reasoning in general, while the remaining percentage may be attributed to language understanding. This pattern may not hold for GPT-4 and Claude-3-opus, as the percentage of consistent incorrect questions falls below 40\%, suggesting that these models may exhibit better commonsense understanding but face challenges in Thai language understanding. The full details can be found in appendix \ref{sec:consistence}.
\section{Conclusion}
In conclusion, the Thai Winograd Schemas benchmark represents a noteworthy step in evaluating commonsense reasoning capabilities within the Thai language context. This novel dataset, meticulously developed and validated by native speakers and professional translators, aims to preserve linguistic and cultural nuances unique to Thai. The comprehensive evaluation of state-of-the-art language models, including GPT-4, Claude-3 variants, Typhoon, and Command R+ on both English and Thai versions of the Winograd Schema Challenge offers insights into their cross-lingual performance. The observed performance drop in Thai suggests challenges these models may face in handling low-resource languages, indicating a need for further research and development in multilingual natural language processing.
\section*{Limitations}
While cultural nuances are preserved as much as possible during the translation process, it is acknowledged that complete preservation is not always achievable due to differences between Thai and English. This linguistic difference may also contribute to the slightly lower performance observed for the Thai human baseline compared to its English counterpart. Direct evaluation on large language models like GPT-4 and Claude-3 cannot be performed due to lack of access and insufficient computational power. Therefore, an alternative approach using prompt-based evaluation is adopted.
\section*{Acknowledgments}
Thank you to the two translators, Chanikarn Inthongpan and Korakoch Rienmek, who worked on translating these Winograd schemas. It was challenging to find individuals skilled enough for this specific task, which demands a deep understanding of language. I also thank the validators Sakrapee Namsak and Chonnawit Khumchoo for their careful review and feedback that greatly contributed to the quality and accuracy of the final dataset. Finally, thank you to Professor Ernest Davis, Professor Chinmay Hegde, and Vid Kocijan for their helpful feedback and support throughout this process.
\bibliography{custom}

\begin{thebibliography}{33}
\providecommand{\natexlab}[1]{#1}

\bibitem[{Achiam et~al.(2023)Achiam, Adler, Agarwal, Ahmad, Akkaya, Aleman, Almeida, Altenschmidt, Altman, Anadkat et~al.}]{achiam2023gpt}
Josh Achiam, Steven Adler, Sandhini Agarwal, Lama Ahmad, Ilge Akkaya, Florencia~Leoni Aleman, Diogo Almeida, Janko Altenschmidt, Sam Altman, Shyamal Anadkat, et~al. 2023.
\newblock Gpt-4 technical report.
\newblock \emph{arXiv preprint arXiv:2303.08774}.

\bibitem[{Amsili and Seminck(2017)}]{amsili2017google}
Pascal Amsili and Olga Seminck. 2017.
\newblock A google-proof collection of french winograd schemas.
\newblock In \emph{The 2nd Workshop on Coreference Resolution Beyond OntoNotes (CORBON 2017), co-located with EACL 2017}, pages 24--29.

\bibitem[{Anthropic(2024{\natexlab{a}})}]{claude3}
Anthropic. 2024{\natexlab{a}}.
\newblock Introducing the next generation of claude.
\newblock \url{https://www.anthropic.com/news/claude-3-family}.
\newblock Accessed: 2024-05-17.

\bibitem[{Anthropic(2024{\natexlab{b}})}]{anthropicSystemPrompts}
Anthropic. 2024{\natexlab{b}}.
\newblock System prompts documentation.
\newblock \url{https://docs.anthropic.com/en/docs/system-prompts}.
\newblock Accessed: 2024-05-17.

\bibitem[{Bernard and Han(2020)}]{bernard-han-2020-mandarinograd}
Timoth{\'e}e Bernard and Ting Han. 2020.
\newblock \href {https://aclanthology.org/2020.lrec-1.3} {{M}andarinograd: A {C}hinese collection of {W}inograd schemas}.
\newblock In \emph{Proceedings of the Twelfth Language Resources and Evaluation Conference}, pages 21--26, Marseille, France. European Language Resources Association.

\bibitem[{Brown et~al.(2020)Brown, Mann, Ryder, Subbiah, Kaplan, Dhariwal, Neelakantan, Shyam, Sastry, Askell et~al.}]{brown2020language}
Tom Brown, Benjamin Mann, Nick Ryder, Melanie Subbiah, Jared~D Kaplan, Prafulla Dhariwal, Arvind Neelakantan, Pranav Shyam, Girish Sastry, Amanda Askell, et~al. 2020.
\newblock Language models are few-shot learners.
\newblock \emph{Advances in neural information processing systems}, 33:1877--1901.

\bibitem[{Coda-Forno et~al.(2024)Coda-Forno, Binz, Wang, and Schulz}]{pmlr-v235-coda-forno24a}
Julian Coda-Forno, Marcel Binz, Jane~X Wang, and Eric Schulz. 2024.
\newblock \href {https://proceedings.mlr.press/v235/coda-forno24a.html} {{C}og{B}ench: a large language model walks into a psychology lab}.
\newblock In \emph{Proceedings of the 41st International Conference on Machine Learning}, volume 235 of \emph{Proceedings of Machine Learning Research}, pages 9076--9108. PMLR.

\bibitem[{Cohere(2024)}]{cohere2024commandrplus}
Cohere. 2024.
\newblock Command r+ with microsoft azure.
\newblock \url{https://cohere.com/blog/command-r-plus-microsoft-azure}.
\newblock Accessed: 2024-12-13.

\bibitem[{Conneau et~al.(2018)Conneau, Rinott, Lample, Williams, Bowman, Schwenk, and Stoyanov}]{conneau2018xnli}
Alexis Conneau, Ruty Rinott, Guillaume Lample, Adina Williams, Samuel~R. Bowman, Holger Schwenk, and Veselin Stoyanov. 2018.
\newblock Xnli: Evaluating cross-lingual sentence representations.
\newblock In \emph{Proceedings of the 2018 Conference on Empirical Methods in Natural Language Processing}. Association for Computational Linguistics.

\bibitem[{Davis(2023)}]{davis2023benchmarks}
Ernest Davis. 2023.
\newblock Benchmarks for automated commonsense reasoning: A survey.
\newblock \emph{ACM Computing Surveys}, 56(4):1--41.

\bibitem[{Davis et~al.(2016)Davis, Morgenstern, and Ortiz}]{davis2016human}
Ernest Davis, Leora Morgenstern, and Charles Ortiz. 2016.
\newblock Human tests of materials for the winograd schema challenge 2016.
\newblock \emph{URL: https://cs. nyu. edu/faculty/davise/papers/WS2016SubjectTests. pdf}.

\bibitem[{De~Melo et~al.(2019)De~Melo, Imaizumi, and Cozman}]{de2019winograd}
Gabriela De~Melo, Vinicius Imaizumi, and F{\'a}bio Cozman. 2019.
\newblock Winograd schemas in portuguese.
\newblock In \emph{Anais do XVI Encontro Nacional de Intelig{\^e}ncia Artificial e Computacional}, pages 787--798. SBC.

\bibitem[{Dubey et~al.(2024)Dubey, Jauhri, Pandey, Kadian, Al-Dahle, Letman, Mathur, Schelten, Yang, Fan et~al.}]{dubey2024llama}
Abhimanyu Dubey, Abhinav Jauhri, Abhinav Pandey, Abhishek Kadian, Ahmad Al-Dahle, Aiesha Letman, Akhil Mathur, Alan Schelten, Amy Yang, Angela Fan, et~al. 2024.
\newblock The llama 3 herd of models.
\newblock \emph{arXiv preprint arXiv:2407.21783}.

\bibitem[{Elazar et~al.(2021)Elazar, Zhang, Goldberg, and Roth}]{elazar-etal-2021-back}
Yanai Elazar, Hongming Zhang, Yoav Goldberg, and Dan Roth. 2021.
\newblock \href {https://doi.org/10.18653/v1/2021.emnlp-main.819} {Back to square one: Artifact detection, training and commonsense disentanglement in the {W}inograd schema}.
\newblock In \emph{Proceedings of the 2021 Conference on Empirical Methods in Natural Language Processing}, pages 10486--10500, Online and Punta Cana, Dominican Republic. Association for Computational Linguistics.

\bibitem[{Kocijan et~al.(2023)Kocijan, Davis, Lukasiewicz, Marcus, and Morgenstern}]{kocijan2023defeat}
Vid Kocijan, Ernest Davis, Thomas Lukasiewicz, Gary Marcus, and Leora Morgenstern. 2023.
\newblock The defeat of the winograd schema challenge.
\newblock \emph{Artificial Intelligence}, page 103971.

\bibitem[{Levesque et~al.(2012)Levesque, Davis, and Morgenstern}]{levesque2012winograd}
Hector Levesque, Ernest Davis, and Leora Morgenstern. 2012.
\newblock The winograd schema challenge.
\newblock In \emph{Thirteenth international conference on the principles of knowledge representation and reasoning}.

\bibitem[{Linzen(2020)}]{linzen-2020-accelerate}
Tal Linzen. 2020.
\newblock \href {https://doi.org/10.18653/v1/2020.acl-main.465} {How can we accelerate progress towards human-like linguistic generalization?}
\newblock In \emph{Proceedings of the 58th Annual Meeting of the Association for Computational Linguistics}, pages 5210--5217, Online. Association for Computational Linguistics.

\bibitem[{Pipatanakul et~al.(2023)Pipatanakul, Jirabovonvisut, Manakul, Sripaisarnmongkol, Patomwong, Chokchainant, and Tharnpipitchai}]{pipatanakul2023typhoon}
Kunat Pipatanakul, Phatrasek Jirabovonvisut, Potsawee Manakul, Sittipong Sripaisarnmongkol, Ruangsak Patomwong, Pathomporn Chokchainant, and Kasima Tharnpipitchai. 2023.
\newblock Typhoon: Thai large language models.
\newblock \emph{arXiv preprint arXiv:2312.13951}.

\bibitem[{Ponti et~al.(2020)Ponti, Glava{\v{s}}, Majewska, Liu, Vuli{\'c}, and Korhonen}]{ponti-etal-2020-xcopa}
Edoardo~Maria Ponti, Goran Glava{\v{s}}, Olga Majewska, Qianchu Liu, Ivan Vuli{\'c}, and Anna Korhonen. 2020.
\newblock \href {https://doi.org/10.18653/v1/2020.emnlp-main.185} {{XCOPA}: A multilingual dataset for causal commonsense reasoning}.
\newblock In \emph{Proceedings of the 2020 Conference on Empirical Methods in Natural Language Processing (EMNLP)}, pages 2362--2376, Online. Association for Computational Linguistics.

\bibitem[{Radford et~al.(2019)Radford, Wu, Child, Luan, Amodei, Sutskever et~al.}]{radford2019language}
Alec Radford, Jeffrey Wu, Rewon Child, David Luan, Dario Amodei, Ilya Sutskever, et~al. 2019.
\newblock Language models are unsupervised multitask learners.
\newblock \emph{OpenAI blog}, 1(8):9.

\bibitem[{Raffel et~al.(2020)Raffel, Shazeer, Roberts, Lee, Narang, Matena, Zhou, Li, and Liu}]{raffel2020exploring}
Colin Raffel, Noam Shazeer, Adam Roberts, Katherine Lee, Sharan Narang, Michael Matena, Yanqi Zhou, Wei Li, and Peter~J Liu. 2020.
\newblock Exploring the limits of transfer learning with a unified text-to-text transformer.
\newblock \emph{Journal of machine learning research}, 21(140):1--67.

\bibitem[{Shavrina et~al.(2020)Shavrina, Fenogenova, Emelyanov, Shevelev, Artemova, Malykh, Mikhailov, Tikhonova, Chertok, and Evlampiev}]{shavrina2020russiansuperglue}
Tatiana Shavrina, Alena Fenogenova, Anton Emelyanov, Denis Shevelev, Ekaterina Artemova, Valentin Malykh, Vladislav Mikhailov, Maria Tikhonova, Andrey Chertok, and Andrey Evlampiev. 2020.
\newblock Russiansuperglue: A russian language understanding evaluation benchmark.
\newblock \emph{arXiv preprint arXiv:2010.15925}.

\bibitem[{Shibata et~al.(2015)Shibata, Kohama, and Kurohashi}]{shibata2015nihongo}
Tomohide Shibata, Shotaro Kohama, and Sadao Kurohashi. 2015.
\newblock Nihongo winograd schema challenge no kouchiku to bunseki.
\newblock \emph{Proceedings of NLP2015}, pages 493--496.

\bibitem[{Shwartz(2024)}]{shwartzWinograd}
Vered Shwartz. 2024.
\newblock Winograd schema challenge datasets.
\newblock \url{https://www.cs.ubc.ca/~vshwartz/resources/winograd_he.html}.
\newblock Accessed: 2024-05-17.

\bibitem[{Su{\'a}rez et~al.(2019)Su{\'a}rez, Sagot, and Romary}]{suarez2019asynchronous}
Pedro Javier~Ortiz Su{\'a}rez, Beno{\^\i}t Sagot, and Laurent Romary. 2019.
\newblock Asynchronous pipeline for processing huge corpora on medium to low resource infrastructures.
\newblock In \emph{7th Workshop on the Challenges in the Management of Large Corpora (CMLC-7)}. Leibniz-Institut f{\"u}r Deutsche Sprache.

\bibitem[{Tanaka et~al.(2013{\natexlab{a}})Tanaka, Rzepka, and Katajima}]{tanaka2024translationpreserving}
Soichiro Tanaka, Rafal Rzepka, and Shiho Katajima. 2013{\natexlab{a}}.
\newblock Translation preserving english names.
\newblock \url{http://arakilab.media.eng.hokudai.ac.jp/~kabura/collection_katakana.html}.
\newblock Translation preserving English names, PDF and HTML formats.

\bibitem[{Tanaka et~al.(2013{\natexlab{b}})Tanaka, Rzepka, and Katajima}]{tanaka2024translations}
Soichiro Tanaka, Rafal Rzepka, and Shiho Katajima. 2013{\natexlab{b}}.
\newblock Translations into japanese.
\newblock \url{http://arakilab.media.eng.hokudai.ac.jp/~kabura/collection_ja.html}.
\newblock Translation changing English names to Japanese, PDF and HTML formats.

\bibitem[{Trinh and Le(2018)}]{trinh2018simple}
Trieu~H Trinh and Quoc~V Le. 2018.
\newblock A simple method for commonsense reasoning.
\newblock \emph{arXiv preprint arXiv:1806.02847}.

\bibitem[{Vad{\'a}sz and Ligeti-Nagy(2022)}]{vadasz2022winograd}
No{\'e}mi Vad{\'a}sz and No{\'e}mi Ligeti-Nagy. 2022.
\newblock Winograd schemata and other datasets for anaphora resolution in hungarian.
\newblock \emph{Acta Linguistica Academica}, 69(4):564--580.

\bibitem[{Wang et~al.(2019)Wang, Pruksachatkun, Nangia, Singh, Michael, Hill, Levy, and Bowman}]{wang2019superglue}
Alex Wang, Yada Pruksachatkun, Nikita Nangia, Amanpreet Singh, Julian Michael, Felix Hill, Omer Levy, and Samuel Bowman. 2019.
\newblock Superglue: A stickier benchmark for general-purpose language understanding systems.
\newblock \emph{Advances in neural information processing systems}, 32.

\bibitem[{Wang et~al.(2018)Wang, Singh, Michael, Hill, Levy, and Bowman}]{wang-etal-2018-glue}
Alex Wang, Amanpreet Singh, Julian Michael, Felix Hill, Omer Levy, and Samuel Bowman. 2018.
\newblock \href {https://doi.org/10.18653/v1/W18-5446} {{GLUE}: A multi-task benchmark and analysis platform for natural language understanding}.
\newblock In \emph{Proceedings of the 2018 {EMNLP} Workshop {B}lackbox{NLP}: Analyzing and Interpreting Neural Networks for {NLP}}, pages 353--355, Brussels, Belgium. Association for Computational Linguistics.

\bibitem[{Winograd(1972)}]{winograd1972understanding}
Terry Winograd. 1972.
\newblock Understanding natural language.
\newblock \emph{Cognitive psychology}, 3(1):1--191.

\bibitem[{Zhang et~al.(2024)Zhang, Aljunied, Gao, Chia, and Bing}]{zhang2024m3exam}
Wenxuan Zhang, Mahani Aljunied, Chang Gao, Yew~Ken Chia, and Lidong Bing. 2024.
\newblock M3exam: A multilingual, multimodal, multilevel benchmark for examining large language models.
\newblock \emph{Advances in Neural Information Processing Systems}, 36.

\end{thebibliography}

\appendix

\section{Prompt Evaluation}
\label{sec:prompteval}
To ensure consistency and reproducibility in evaluating each model, specific settings were implemented alongside the prompt structure, as illustrated in Figure~\ref{fig:evaluation-method}. 

The prompt follows a consistent pattern to ensure clarity and replicability across evaluations:
\begin{enumerate}
\item \textbf{Sentence with Pronoun}: The main sentence is provided, with the ambiguous pronoun enclosed in asterisks (**), followed by a newline character (\textbackslash n).
\item \textbf{Snippet:}: A shortened snippet containing the pronoun and its immediate context is included after the sentence, labeled as “Snippet: ”, followed by a newline character (\textbackslash n).
\item \textbf{Options:}: The candidate antecedents for the pronoun are listed after “Options:”, separated by newline characters (\textbackslash n), with no newline after the final option.
\end{enumerate}

This process was executed through the model API, where each model was prompted with the designed prompts, and the answers were obtained directly from the model’s output. This ensured that all evaluations followed the same method and settings across models.
\begin{figure}
    \centering
    \begin{tcolorbox}[colback=white, colframe=black, width=\columnwidth]
        \textbf{System Prompt:}  
        \begin{quote}
            "You will be provided with a sentence and a snippet containing a pronoun enclosed in asterisks (**). Your task is to determine the correct referent of the pronoun from the given options. Respond only with one of the provided choices, exactly as it is written. For example, if the options are ‘The city councilmen’ and ‘The demonstrators’, respond only with ‘The city councilmen’ or ‘The demonstrators’."
        \end{quote}
        
        \textbf{User Prompt:}
        \begin{quote}
        "The city councilmen refused the demonstrators a permit because **they** feared violence.\textbackslash nSnippet: **they** feared violence\textbackslash nOptions:\textbackslash nThe city councilmen\textbackslash nThe demonstrators"
        \end{quote}        
        \textbf{Accepted Answer:}  
        \begin{quote}
            "The city councilmen"
        \end{quote}
        
        \textbf{Unaccepted Answer:}  
        \begin{quote}
            "The answer is The city councilmen"
        \end{quote}
    \end{tcolorbox}
    \caption{An example of the prompt evaluation method, detailing the system prompt, user prompt, and expected answer. Only exact matches like "The city councilmen" were considered correct, while responses such as "The answer is The city councilmen" were not accepted, ensuring consistent and reproducible evaluations.}
    \label{fig:evaluation-method}
\end{figure}
The code used to reproduce all of our experimental results is available at the following GitHub repository: \url{https://github.com/PhakphumAdev/Thai-Winograd}.

\section{Models Used in the Study}
\label{sec:testedmodels}
\subsection*{Typhoon}
The Thai large language model, named Typhoon, was initially built on the Mistral-7B architecture \cite{pipatanakul2023typhoon}. Typhoon was pre-trained on the MC4 \cite{raffel2020exploring} and OSCAR \cite{suarez2019asynchronous} datasets, which include approximately 3 TB of Thai text. In this experiment, the specific model tested was Typhoon-Instruct, as documented on the Open Typhoon website\footnote{\url{https://docs.opentyphoon.ai}}. This version of Typhoon, tested in September 2024, is based on the LLaMA 3 8B architecture \cite{dubey2024llama}. It was released on September 5, 2024, and is distinct from versions built on the Mistral-7B architecture.
\subsection*{Claude}
Anthropic’s large language model \cite{claude3}, designed to be a helpful and honest assistant, has its model architecture details kept private. In this experiment, all the Claude-3 models were tested to compare performance. The specific models tested were claude-3-haiku-20240307, claude-3-sonnet-20240229, and claude-3-opus-20240229.
\subsection*{GPT}
GPT or Generative Pre-trained Transformer is one of the most popular and powerful large language models available. While the specific details of GPT-4's architecture are not publicly disclosed, it can be assumed that it represents an advancement over GPT-3. GPT-3 is an autoregressive, decoder-only model with 175 billion parameters \cite{brown2020language}. Although the technical report does not specify its performance in Thai specifically, it asserts that GPT-4 surpasses GPT-3 in the MMLU task for the Thai language \cite{achiam2023gpt}. In this experiment, gpt-4-0613 and gpt-3.5-turbo-0125 were tested.
\subsection*{C4AI Command R+}
C4AI Command R+ is a 104-billion parameter model with advanced capabilities, including Retrieval Augmented Generation (RAG) and multi-step tool use for automating complex tasks \cite{cohere2024commandrplus}. Optimized for reasoning, summarization, and question answering, it excels across various use cases. In this experiment, command-r-plus-08-2024 were tested.

\section{Consistency of errors in English LLMs}
\label{sec:consistence}
\begin{table*}[htbp]
\centering
\begin{tabular}{lccc}
\hline
Model         & \% for incorrect overlapping answers (Thai) & \% for incorrect overlapping answers (English) \\ \hline
Typhoon       & 56.80\%             & 60.17\% \\
Claude-3-Haiku         & 51.88\%             & 67.65\% \\
Claude-3-Sonnet      & 33.68\%           & 58.18\% \\
Claude-3-Opus     & 15.52\%            & 42.86\% \\
GPT-3.5         & 34.59\%             & 55.42\% \\
GPT-4       & 8.96\%             & 35.29\% \\ 
Command-r-plus & 22.94\%             & 67.57\% \\
\hline
\end{tabular}
\caption{Consistency of errors}
\label{tab:error}
\end{table*}

\end{document}